%% file: aaai24.tex
\documentclass[letterpaper]{article} % DO NOT CHANGE THIS
\usepackage[bottom]{footmisc}
\usepackage{aaai24}  % DO NOT CHANGE THIS
\usepackage{times}  % DO NOT CHANGE THIS
\usepackage{helvet}  % DO NOT CHANGE THIS
\usepackage{courier}  % DO NOT CHANGE THIS
\usepackage[hyphens]{url}  % DO NOT CHANGE THIS
\usepackage{graphicx} % DO NOT CHANGE THIS
\usepackage{natbib}  % DO NOT CHANGE THIS AND DO NOT ADD ANY OPTIONS TO IT
\usepackage{caption} % DO NOT CHANGE THIS AND DO NOT ADD ANY OPTIONS TO IT
\usepackage{algorithm}
\usepackage{algorithmic}

\usepackage{multirow}

\usepackage{newfloat}
\usepackage{listings}
\DeclareCaptionStyle{ruled}{labelfont=normalfont,labelsep=colon,strut=off} % DO NOT CHANGE THIS
\floatstyle{ruled}
\newfloat{listing}{tb}{lst}{}
\floatname{listing}{Listing}
\newcommand{\ie}{\textit{i.e.}}
\usepackage{amsmath}
\usepackage{amssymb}
\usepackage{amsfonts}

\nocopyright 

\baselinestretch
\clearpage
\title{Product-Level Try-on: Characteristics-preserving Try-on with Realistic Clothes Shading and Wrinkles}
\author {
    Yanlong Zang\textsuperscript{\rm 1}$^*$
    Han Yang\textsuperscript{\rm 2}$^*$
    Jiaxu Miao\textsuperscript{\rm 1}
    Yi Yang\textsuperscript{\rm 1}$^\dag$
}
\affiliations {
    \textbf{\textsuperscript{\rm 1}Zhejiang University\quad \textsuperscript{\rm 2}ETH Zurich} \\
    \textbf{yanlongzang@gmail.com\quad hanyang@ethz.ch\quad jiaxu.miao@yahoo.com \quad yangyics@zju.edu.cn}
}
\usepackage{bibentry}
\begin{document}

\maketitle

\begin{abstract}
Image-based virtual try-on systems for fitting new in-shop garments into human portraits have attracted increasing research attention. An ideal pipeline should not only preserve the in-shop clothes static characteristics (e.g. textures, logos, embroideries) in the generated images but also generate dynamic features (e.g. shadow, folds) that change according to the model pose and the environmental ambiance.

Previous works fail specifically in generating dynamic features, as they preserve the warped in-shop clothes trivially with predicted an alpha mask by composition. To break the dilemma of over-preserving and textures losses, we propose a novel diffusion-based Product-level virtual try-on pipeline, \ie PLTON, which can preserve the fine details of logos and embroideries while producing realistic clothes shading and wrinkles.\footnote{* means equal contribution. $\dag$ indicates corresponding author.} The main insights are in three folds: 
\textbf{1)} Adaptive Dynamic Rendering: We take a pre-trained diffusion model as a generative prior and tame it with image features, training a dynamic extractor from scratch to generate dynamic tokens that preserve high-fidelity semantic information. Due to the strong generative power of the diffusion prior, we can generate realistic clothes shadows and wrinkles. \textbf{2)} Static Characteristics Transformation: High-frequency Map (HF-Map) is our fundamental insight for static representation. PLTON first deforms the in-shop clothes to the target model pose by a traditional warping network, then we use a high-pass filter to extract an example HF-Map to preserve the static features of the clothes. The HF-Map is then fed into our static extractor to generate modulation maps, which are injected into the fixed U-net structure to synthesize the final result. \textbf{3)} To further enhance retention, the Two-stage Blended Denoising method is proposed to guide the diffusion process toward the correct spatial layout and color.

PLTON is finetuned only with our collected small-size try-on dataset. Extensive quantitative and qualitative experiments on $1024\times 768$ datasets demonstrate the superiority of our framework in mimicking real clothes dynamics.
\end{abstract}

\input{sections/introduction}

\input{sections/related_works}

\input{sections/methods}

\input{sections/experiments}

\input{sections/conclusion}

\bibliography{aaai24}

\end{document}

%% file: sections/introduction.tex
\section{Introduction}

With the continuous development of generative models, virtual try-on has become an increasingly popular topic. While GAN-based methods ~\cite{RT-VTON, VITON-HD, HR-VTON} can generate high-resolution (1024×768) images that preserve nearly all clothes static characteristics (e.g. textures, logos) and dynamic features (e.g., shadings, folds) of in-shop clothes, questions arise about whether these results meet our needs. \textbf{1)} For instance, in a fashion editorial style, consider the scenario where different models wear the same clothes in various environments. Should the clothes' dynamic features always maintain the same characteristics as in-shop garments? \textbf{2)} Can we take advantage of large models with greater generative capabilities than GANs to accurately generate natural and dynamic clothes features related to model pose and environmental atmosphere? Our investigation into previous virtual try-on methods and the application of large models in image generation have the following several key findings.

\begin{figure}[t]
\centering
\includegraphics[width=0.9\columnwidth]{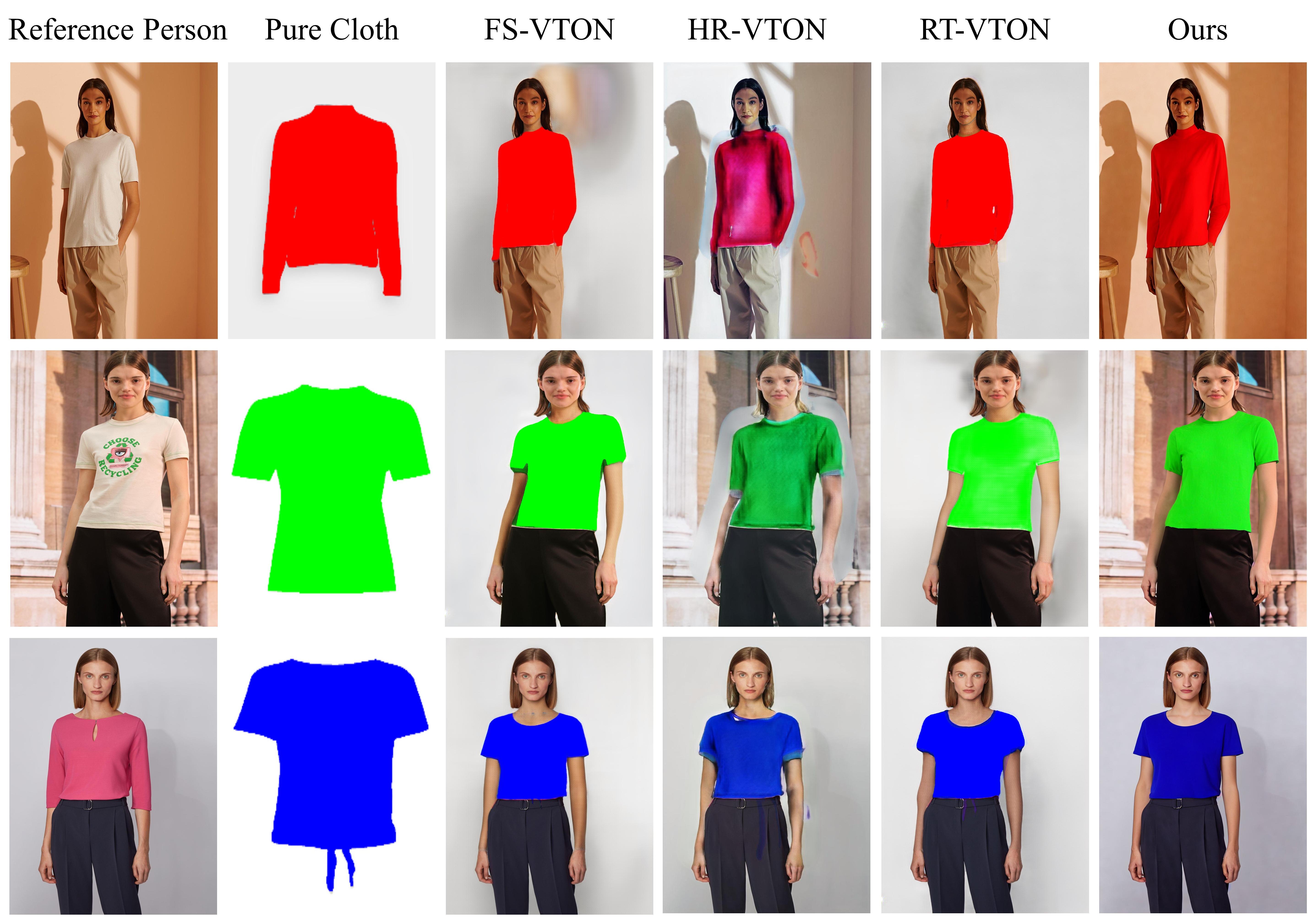} % Reduce the figure size so that it is slightly narrower than the column. Don't use precise values for figure width.This setup will avoid overfull boxes.
\caption{Visual comparison of PLTON and other four traditional virtual try-on algorithms in generating clothes shadows and folds. To eliminate the influence of the original clothes' dynamic features, we fill the clothes with three solid colors: red, green, and blue (from top to bottom).}
\label{fig:introduction}
\vspace{-15pt}
\end{figure}

Firstly, we have observed that traditional virtual try-on pipelines involve inputting warped clothes into the generative model. However, improving the conventional virtual try-on pipeline is challenging due to the difficulty in directly disentangling typical clothes static characteristics and dynamic features. Follow-up methods such as \cite{FS-VTON, PFAFN, RT-VTON, CP-VTON, CP-VTON+} aim to preserve the static characteristics of deformed in-shop clothes by generating a compositional mask. Since most in-shop clothes are stereoscopic, the unprocessed dynamic features (e.g., shadings and folds) are mis-preserved in the final synthesis causing incoherent human-background lighting conditions. Moreover, when we eliminate the shadow and folds of in-shop clothes (filling the in-shop clothes with three solid colors: red, green, and blue), as shown in Figure \ref{fig:introduction}, traditional virtual try-on algorithms~\cite{RT-VTON, FS-VTON, HR-VTON} can hardly generate natural dynamic shadings and folds.

Secondly, while the diffusion model exhibits stronger generative capabilities, generating high-resolution images often requires expensive data and computing resources. However, high-resolution virtual try-on data is not always accessible, and the current open-source data set is generally about 10k, training a photo-realistic try-on pipeline with limited try-on pair data is a valuable topic.

% To design a lightweight try-on pipeline with limited data and computational resources, we compress the high-level semantics of target clothes, leverage the solid generative abilities of pre-trained diffusion prior, and decouple the generation process into two semantic domains as static preserving and dynamic rendering. Without explicit 3D modeling, static features such as logos and embroideries are represented by the high-frequency map (HF-Map), while dynamic features such as shading and wrinkles are integrated implicitly in diffusion prior. This way, we can largely generate clothes-specific fine details without losing many clothes characteristics.

\begin{figure*}[thb]
\begin{center}
\includegraphics[width=0.95\linewidth]{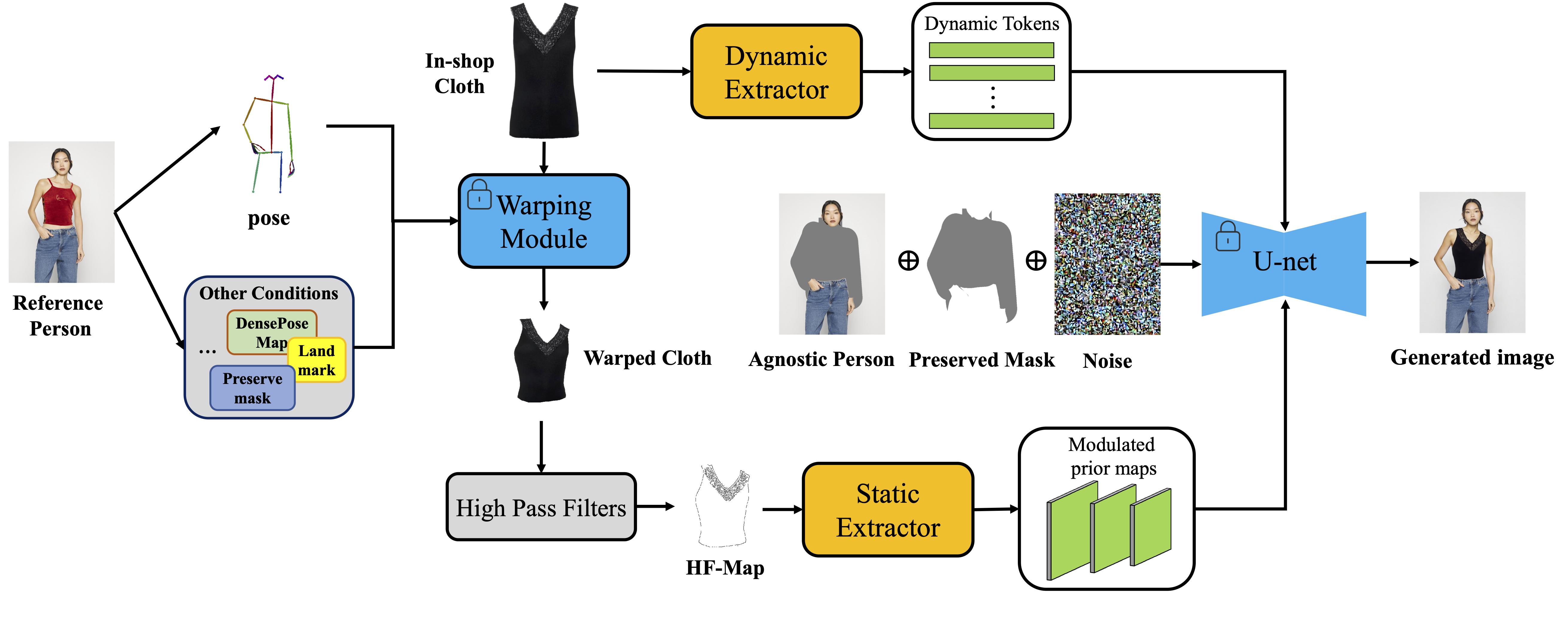}
\vspace{-10pt}
\end{center}
    \caption{A schematic of PLTON. We utilize the warping module to deform the in-shop clothes using pose ~\cite{openpose} and other conditions (e.g. densepose ~\cite{densepose} segmentation ~\cite{graphonomy,atr}). Firstly, we apply High Pass Filters to the warped cloth to extract high-frequency features of the clothes. Then, we employ a Static Extractor to extract modulated prior maps from the HF-Map. Subsequently, the Dynamic Extractor is utilized to extract the dynamic features of the in-shop cloth, generating dynamic tokens. Finally, the dynamic tokens and modulated prior maps are input into a fixed pre-trained diffusion model, which produces the final output. The terms "Locked" and "Lockless" represent frozen and learnable parameters, respectively.}
   
\label{fig:pipeline}
\vspace{-12pt}
\end{figure*}

We propose the product-level try-on (PLTON) to address the above challenges. Unlike previous traditional try-on methods, PLTON is able to preserve static in-shop clothes details such as textures, logos, and embroideries, while generating realistic clothes shadows and folds with limited data. Specifically, we decouple the traditional one-stage clothes synthesis process into Adaptive Dynamic Rendering and Static Characteristics Transformation. We use a dynamic extractor to extract dynamic tokens on compressed in-shop clothes and use a static extractor to extract modulated prior maps from the high-frequency map (HF-Map) as a supplement. We then inject dynamic tokens and modulated prior maps into a fixed pre-trained diffusion model to guide image generation. In order to reduce the information loss caused by compressing in-shop clothes in the dynamic extractor, we propose two-stage blended denoising, which can simultaneously solve the problem of repeated patterns when generating high-resolution images, and guide the diffusion process toward the correct spatial layout and colors, resulting in more accurate and precise outputs.

PLTON was trained on a small try-on dataset (less than 20k) collected from the Internet and achieved SOTA results on a high-resolution dataset using only one A40, illustrating the effectiveness and training efficiency of our PLTON.

% We have trained PLTON on a small try-on dataset (less than 20k) and achieved SOTA results on high-resolution datasets only using one single A40, illustrating the effectiveness and training efficiency of our PLTON.

%% file: sections/related_works.tex
\section{Related Works}

\noindent \textbf{Image-based Virtual Try-on.}\quad
The goal of image-based virtual tyr-on is to naturally and realistically transfer in-shop clothes to a reference person. Image-based virtual try-on can be divided into two settings: model-to-model try-on such as \cite{M2e-tryon,MG-VTON,PASTA-GAN,Outfit-VITON} and cloth-to-model such as \cite{VITON,CP-VTON,ACGPN,PFAFN,FS-VTON,DAFlow,ZFlow,Clothflow,GPflow}. Our primary focus is on the cloth-to-model setting.

The recent virtual try-on algorithms generally comprise a warping module and a fusion module. The fusion module uses warped clothes obtained by the warping module and other conditions (e.g., after-try-on human parsing) to generate the try-on results. There are two mainstream methods of the fusion module: 1) ~\cite{CP-VTON, CP-VTON+, PFAFN, HR-VTON} preserve the warped in-shop clothes trivially with predicting an alpha mask by composition, but easily results in unrealistic images when the in-shop garments dynamic features do not match the background and lighting conditions of the reference person. 2) ~\cite{VITON-HD, HR-VTON} proposed a GAN-based generator to generate the clothes dynamic features, but the static characteristics of in-shop garments will be blurred when generating high-resolution images. In this paper, we target to accurately capture the in-shop clothes' static characteristics and naturally generate clothes dynamic features.

\noindent \textbf{Diffusion Probabilistic Model.}\quad
The diffusion probability model, including forward and inverse processes, was introduced in ~\cite{DM2015}. The model was used in image generation, with DDPM ~\cite{DDPM} being the first to synthesize high-quality images. The inverse process in image generation involves converting raw images into Gaussian distributions, adding random Gaussian noise, and then recovering the raw image through several denoising steps. DDIM ~\cite{DDIM} improved the reverse process to reduce denoising steps and increased sampling speed. The diffusion model has shown more excellent generative capabilities than the long-dominant GAN~\cite{GAN} in many challenging image synthesis tasks ~\cite{Segdiff,Label-efficient-diffusion,brempong2022denoising, cai2020learning}.

However, generating high-quality images with diffusion models requires expensive computing resources and data. For example, generating images with given prompts generally requires millions of training data~\cite{DALLE-2,Imagen}. Furthermore, generating images with given images is more challenging. Even by fine-tuning a large diffusion model, millions of data are required. For example, ~\cite{PBE} used 1.9 million data and trained 64 v100s for seven days. Collecting paired image data is difficult in virtual try-on, and the largest open-source virtual try-on dataset VITON ~\cite{VITON} has only 14,221 images. To address this challenge, we propose a novel approach to fine-tune large models by fixing the network and controlling the bootstrap only with additional conditions.
% \noindent \textbf{Personalization and Control of Pretrained Diffusion Model.}\quad
% The current state-of-the-art diffusion models use prompts to control image generation, but sometimes it is difficult for people to describe what they want to generate. ~\cite{ControlNet} proposed a general module that can generate an image based on coarse-grained image conditions (e.g. pose, segmentation). Additionally, ~\cite{PBE} provides fine-grained control over changing image content based on example image semantics, offering a more intuitive approach to simplify fine-grained image editing for novices. Despite these impressive results, accurately preserving the details of exemplar images under high resolution remains a challenge.

\noindent \textbf{Clothes Composition.}\quad
we reiterate the pros and cons of the conventional design of composition-based try-on. Traditional GAN-based pipelines employ a split-transform-merge pipeline with a final alpha composition to fuse all image components to generate a synthetic clothed person. As proposed by VITON~\cite{VITON}, alpha masks help composite deformed clothing images with rendered coarse images. Due to the randomness of logos and embroidery, it is almost impossible to generate the full features of the target clothes without a synthetic mask. This acknowledgment seems to dominate the current try-in pipeline design, severely hindering the development of the field. As shown in figure ~\ref{fig:introduction}, it is incredibly challenging to preserve the fine details of the target clothing while generating coherent clothes shadows and wrinkles on the reference person. Our primary motivation is to break this dilemma and design a new paradigm to replace the traditional design of a product-level try-on application.

%% file: sections/methods.tex
\section{Methods}
Our method aims to transfer the in-shop clothes to the reference model vividly and naturally. The generated image showcases the clothes' static characteristics, such as logos and embroideries, while incorporating dynamic features that change with the environment and reference person, such as shadows and folds. To achieve this goal, we fine-tune a large diffusion model based on stable diffusion ~\cite{Stable-diffusion} on our collected dataset. Our approach accepts in-shop and warped clothes signals for retaining static characteristics and rendering dynamic features, resulting in a product-level try-on.

\subsection{Framework Overview}
Our proposed virtual try-on method involves generating images of wearing in-shop clothes using a pre-trained large-scale diffusion model. As shown in Figure ~\ref{fig:pipeline}, PLTON decouples the clothing synthesis process into adaptive dynamic rendering and clothes static and Static Characteristics Transformation. Firstly, we transform the in-shop cloth through the warping module to obtain the warped cloth, and then perform high pass filters on the warped cloth to obtain the high-frequency map (HF-Map). The method used by the warping module here is the style-based appearance flow in ~\cite{FS-VTON}. Then we use dynamic and static extractors to extract dynamic tokens and modulated prior maps from in-shop cloth and HF-Map respectively. Finally, a two-stage blended diffusion method is used for high-resolution inference.

\subsection{Diffusion Prior}
In PLTON, we utilize the current open-source and state-of-the-art stable-diffusion ~\cite{Stable-diffusion} framework as priors, comprising a variational autoencoder (VAE) ~\cite{VAE} and denoiser U-Net ~\cite{U-Net} $\epsilon$. The denoiser operates in a latent space more suitable for likelihood-based generative models than a high-dimensional pixel space. This is because it allows for focusing on critical semantic bits of the data and training in a lower dimensionality, which is more computationally efficient. The first step involves training an autoencoder to compress and reconstruct the original image $x_0$. Subsequently, a modified time-conditioned U-net is trained to iteratively predict the noise corresponding to the latent features at each time step $t \in \{1,...,T\}$. The objective function of U-net is optimized to achieve the desired results:
\begin{equation}
L_{LDM} := \mathbb{E}_{ \varepsilon(x),\epsilon \sim \mathcal{N}(0,1),t}[{\Vert \epsilon - \epsilon_{\theta}(z_t,t,c) \Vert}_2^2],
\end{equation}
where $c$ is the embedding of conditional information. In the previous stable diffusion, the text is passed through the CLIP text encoder and then added to U-Net in the form of cross-attention. In the field of virtual try-on, using prompts to describe the clothes' static characteristics accurately is challenging. To overcome this, we use in-shop clothes images with rich semantics as conditional information embedding.

\subsection{Dynamic Extractor} 
To accurately generate the dynamic features of in-shop clothes (e.g., shading, folds), a simple method is to directly use the in-shop clothes as the conditional input of U-net. However, this approach presents some challenges. Firstly, the process from in-shop clothes to the clothes on models is complex and cross-domain. Secondly, as shown in figure ~\ref{fig:introduction}, if using the warped cloth as input and designing it as a copy-and-paste process, it is essential to note that this approach may result in the loss of the network's ability to generate dynamic features, such as ~\cite{RT-VTON, FS-VTON}.

To address the above issues, we choose a CLIP image encoder to extract the features of in-shop clothes images and decode these features via several additional fully-connected layers. We then inject these features into U-net in the form of cross-attention. We denote the down-sample operation as $\mathcal{D}$. The input in-shop clothes image is represented by $x_{g} \in \mathbb{R}^{h \times w \times c}$, a MLP network $\mathcal{M}(\left .; \theta \right)$ with a set of parameters $\theta$ transforms the embedding extracted by CLIP into another feature map $\mathcal{F}$ with:
\begin{equation}
\mathcal{F} = \mathcal{M}\left( CLIP \left( \mathcal{D} \left( x_{g} \right) \right), \theta \right)
\end{equation}

Nevertheless, the issue of lost information during the compression process has bobbed up while using CLIP. Additionally, obtaining accurate clothes features through training fully connected layers with limited data is challenging. So, we introduce the static extractor stage to auxiliary training. Our approach effectively maintains semantic information while generating dynamic features, even with a small amount of data and computing resources.

\subsection{Static Extractor}
In the Static Extractor, we utilize the high-frequency map (HF-Map) $x_{hf}$ of the warped clothes as additional, conditional information to preserve the clothes' static characteristics. The HF-Map of the warped clothes is first injected into zero-convolution $\mathcal{Z}$ ~\cite{ControlNet} to convert it to the same feature size as a fixed diffusion model $\mathcal{G}_{f}$ input. Then, a trainable diffusion model $\mathcal{G}_{t}$ (encoders and mid-blocks) is cloned from $\mathcal{G}_{f}$ and is used to perform clothes static characteristics extraction. The modulated prior maps are extracted by the encoders in $\mathcal{G}_{t}$ with:
\begin{equation}
y' = \mathcal{Z}\left( 
\mathcal{G}_{t}\left( \widetilde{x}_T + \mathcal{Z}\left(x_{hf}, \theta_{0}\right), \mathcal{F}, \theta_{t} \right),
\theta_{z} \right)
\end{equation}
where $y'$ is the output of the zero-convolution blocks in the trainable diffusion model $\mathcal{G}_{t}$. $\mathcal{F}$ is the dynamic tokens extracted by Dynamic Extractor. Then PLTON uses the fixed diffusion model $\mathcal{G}_{f}$ to fuse modulated prior maps and dynamic tokens with:

\begin{equation}
y = y' + \mathcal{G}_{f} \left( x_{input}, \mathcal{F}, \theta_{f} \right)
\end{equation}

where $y$ becomes the output of the decoder block in the fixed diffusion model $\mathcal{G}_{f}$. Finally, PLTON retains clothes static characteristics and generates dynamic features of clothes successfully.

\begin{algorithm}[tb]
    \caption{Two-stage Blended Denoising}
    \label{alg:algorithm}
    \textbf{Input}: Input Image: $x_0$, Reference Level of the Input Image: $\delta$, Warped Cloth: ${w_0}$, Warped Cloth Mask:${m}$, Reference Level of the Warped Cloth: $\gamma$, The Number of Denoising Steps: $S_{num}$, The list of Denoising Steps: $S_{list}$, Conditions: $c$ \\
    \textbf{Parameter}: PLTON Model: $ \epsilon_\theta$ \\
    \textbf{Output}: Generated image $\widetilde{x}_0$
    \begin{algorithmic}[1] %[1] enables line numbers
        \STATE $T_{num} = S_{num} × \delta$ if $\delta < 1$, else $T_{num}=S_{num}$ \\
        \STATE $T_{start} = S_{num} - T_{num}$ \\
        \STATE $T_{list} = S_{list}\left[T_{start}:\right]$ \\
        \STATE $\eta=0$ \\
        \STATE $\epsilon \sim \mathcal{N}(\textbf{0},\textbf{I})$ \\
        \STATE $\widetilde{x}_t = \sqrt{\alpha_t}x_0 + (1 - \alpha_t)\epsilon $ \\
        \FOR{$t=T_{list}\left[T_{num}\right],...,T_{list}\left[0\right]$}
            % \STATE Do some action.
            \STATE $\widetilde{x}'_{t-1} = \sqrt{\alpha_{t-1}} \left( \frac{x_t-\sqrt{1-\alpha_t}\epsilon_\theta^{t}(\widetilde{x}_t, c )}{\sqrt{\alpha_t}} \right) +$

            \STATE \qquad\quad$\sqrt{1 - \alpha_{t-1} - \sigma_t^2} \cdot \epsilon_\theta^{t}(\widetilde{x}_t,c)+\epsilon_t\sigma_t$
            \IF {$\eta < T_{num} \ast \gamma$}
                \STATE $\epsilon \sim \mathcal{N}(\textbf{0},\textbf{I})$, if $t > 1$, else $\epsilon=\textbf{0}$
                \STATE $w_t = \sqrt{\alpha_t}w + (1 - \alpha_t)\epsilon$
                \STATE $\widetilde{x}_{t-1} = \widetilde{x}'_{t-1} \odot (1 - m) + m \odot w_t$
            \ENDIF
            \STATE $\eta=\eta+1$
        \ENDFOR
        \STATE \textbf{return} $\widetilde{x}_0$
    \end{algorithmic}
\end{algorithm}

\begin{figure*}[t]
\begin{center}
\includegraphics[width=0.85\linewidth]{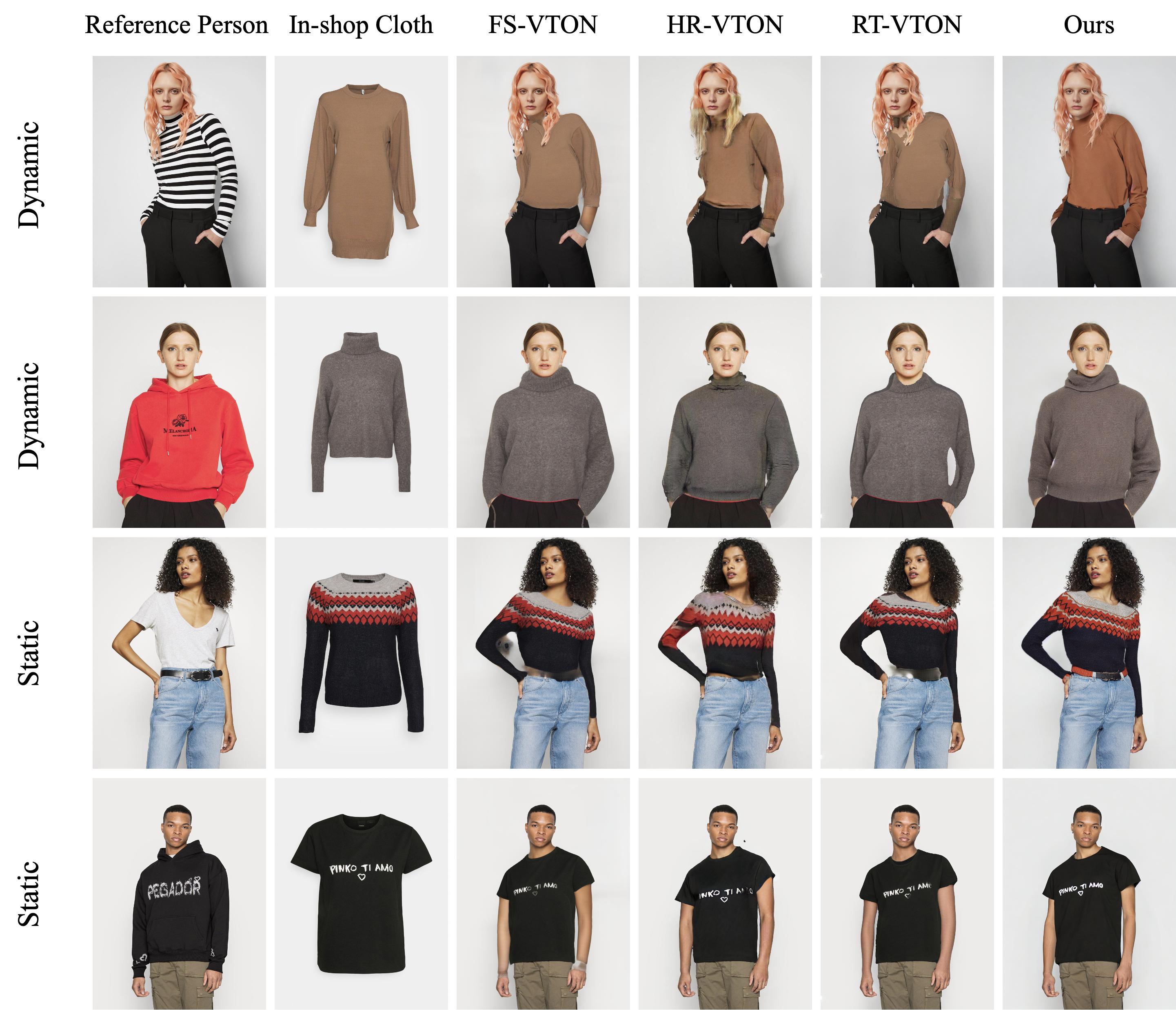}
\vspace{-10pt}
\end{center}
   \caption{The visual comparison of different models (FS-VTON \cite{FS-VTON}, HR-VTON ~\cite{HR-VTON}, RT-VTON ~\cite{RT-VTON} and ours) in the dynamic generation and static characteristics preservation of clothes. 
   }
\label{fig:qualitative}
\vspace{-10pt}
\end{figure*}

\subsection{Two-stage Blended Denoising}
Pasting warped clothes into the generated image may result in the unnatural appearance of clothes boundaries on reference persons. To address this issue, ~\cite{RT-VTON, PFAFN, CP-VTON+} learns an alpha mask to paste the static characteristics of the original in-shop clothes more smoothly. However, as shown in Figure ~\ref{fig:introduction}, this approach sacrifices the model's ability to generate clothes dynamic features.

The static and dynamic extractors discussed in the previous two sections provide precise guidance for the diffusion model. However, due to the limitation of CLIP input size, the high-resolution clothes image is compressed to 224 × 224 resolution, resulting in the information loss of clothes details. 

Secondly, the input image resolution of the diffusion-based methods training is usually 512 × 512. As shown in Figure ~\ref{fig:ablation1}, if we want to get an image with 1024 × 1024 resolution in one step during inference, there will be a repeated patterns problem (see ablation research for details). To address the above problems, we propose two-stage blended denoising, which strengthens the model's retention of static characteristics by using warped clothes with added noise as a guide and allows to adjust the degree of dynamic features generation of clothes through $\delta$ and $\gamma$. In PLTON, we first generate a low-resolution image and then use it as guidance to generate a high-resolution image through image-to-image generation. Both generation processes use Algorithm 1, with $S_{num}$ set to 50. When generating low-resolution images, the input image $x_0$ is noise, and $x_0$ is not used as guidance ($\delta=1$) and the reference degree to warped clothes $\gamma=0.2$. When generating high-resolution images, the input image $x_0$ is the enlarged result of the generated low-resolution image.

%% file: sections/experiments.tex
\section{Experiments}

\subsection{Experiments Setup}
\noindent \textbf{Datasets.}\quad 
We collect a high-resolution (1024 × 768) fashion image dataset from the Internet, consisting of 18,327 frontal view pairs of models and top in-shop clothes images. The dataset is divided into training and test sets, with 15,527 pairs and 2,800 pairs. The test set we have collected is referred to as "TEST1". To further evaluate the generalization ability of PLTON, we perform direct inference on the test set in ~\cite{HR-VTON}, which we call "TEST2". Due to the low resolution of the VITON dataset, we were unable to utilize them in evaluating the generalization ability of our model.

\noindent \textbf{Implementation Details.}\quad 
Our model is implemented in PyTorch and utilizes a single Nvidia A40 GPU for training and high-resolution image inference. The clothes warping module utilizes a StyleGAN-based architecture ~\cite{StyleGAN} in FS-VTON ~\cite{FS-VTON} for the appearance flow strategy. The distillation strategy is not used, and the hyperparameters remain consistent with the FS-VTON open-source code. In order to better extract and mix clothes dynamic tokens and modulated prior maps, we adopt stable-diffusion as our baseline model and utilize the CLIP pre-trained model (ViT-L) as our image encoder in dynamic extractor. We extract the in-shop clothes features from the last hidden state of CLIP as the condition and decode them through 15 fully connected layers. Then we inject extracted features into the diffusion process through cross-attention. Our diffusion priors initialization utilize the publicly released models of ~\cite{PBE} and ~\cite{ControlNet} for the fixed and trainable parts respectively. We train the model using the AdamW optimizer ~\cite{AdamW} with a learning rate of 1e-5, a batch size of 8, and train for 50 epochs. Throughout the training process, we did not use any data augmentation strategy.

\begin{table}[htb]
\begin{center}
  \caption{Quantitative results on two different test sets, TEST1 and TEST2, which are all in $1024\times 768$. We show the FID~\cite{FID} and LPIPS~\cite{lpips}. "*" indicates methods that are only for reference, not the main baselines; discussions are provided in supp.}
  \label{tab:quantitative}
  \begin{tabular}{cccc}
  \hline
  \textbf{Dataset} & \textbf{Method} & \textbf{FID} & \textbf{LPIPS} \\
  \hline
  \multirow{6}{*}{\textbf{TEST1}} & FS-VTON & 9.522 & 0.109 \\
  & HR-VTON & 11.852 & 0.146 \\
  & RT-VTON & 9.051 & 0.116 \\
  & DAFlow$^*$ & 12.110 & 0.113 \\
  & PFAFN$^*$ & 9.892 & 0.114 \\
  & Ours & \textbf{8.394} & 0.113 \\
  \hline
  \hline
  \multirow{6}{*}{\textbf{TEST2}} & FS-VTON & 11.803 & 0.118 \\
  & HR-VTON & 14.684 & 0.122 \\
  & RT-VTON & 11.471 & 0.132 \\
  & DAFlow$^*$ & 15.919 & 0.169 \\
  & PFAFN$^*$ & 12.462 & 0.125 \\
  & Ours & \textbf{11.321} & 0.129 \\
  \hline
  % \bottomrule
\end{tabular}
\end{center}
\vspace{-20pt}
\end{table}

\subsection{Qualitative Results}
In the field of virtual try-on, we present a comparative analysis of our method with several current state-of-the-art baselines, including flow-based FS-VTON~\cite{FS-VTON}, semantic-based methods RT-VTON~\cite{RT-VTON} and HR-VTON~\cite{HR-VTON}. The distillation trick is not applied to all methods for fair comparison. We leverage publicly available open-source code for training or fine-tuning on our dataset. Our visuals on TEST1 are shown in Figure ~\ref{fig:qualitative}, where our method produces more realistic images than baselines. Our model not only well preserves the static characteristics (e.g. textures, logos, and embroideries) of the target clothes but also naturally generates the dynamic features of the clothes (e.g. folds and shadows). Visual comparisons on TEST2 are provided in supp.

\noindent \textbf{Effectiveness of Dynamic Extractor.}\quad 
Dynamic Extractor is crucial in enhancing the realism of diffusion model-generated dynamic features. In contrast to the traditional virtual try-on algorithms FS-VTON~\cite{FS-VTON} and RT-VTON~\cite{RT-VTON}, which rely on learning an alpha mask to composite the warped cloth and generated image, the Dynamic Extractor module guides the diffusion model to produce more natural-looking images. The resulting images exhibit more realistic lighting, shadows, and folds, as exemplified by the sleeves and waist of the clothes (row 1 in Figure ~\ref{fig:qualitative}) and the high collar area (row 1 in Figure ~\ref{fig:qualitative}). As shown in Figure ~\ref{fig:qualitative}, while HR-VTON leverages GAN to generate clothes dynamic features, the resulting shadows and folds look dirty. By leveraging the guidance provided by the Dynamic Extractor module, PLTON leverages the strengths of the large model to produce more reliable, natural, and realistic clothing dynamic features.

\noindent \textbf{Effectiveness of Static Extractor.}\quad
The Static Extractor is designed to enhance the diffusion model's ability to preserve clothes static characteristics by introducing additional conditional information (e.g. high-frequency map of warped clothes). While FS-VTON and RT-VTON rely on learning the composition mask to paste distorted clothes static features back into the generated image. However, as shown in  Figure ~\ref{fig:qualitative} (row.3, row.4), these methods can only copy and paste part of the textures when the alpha mask learning is inaccurate. Similarly, when the warped clothes cannot be well aligned with the model, only part of the static features of the garment can be preserved. Additionally, HR-VTON, which totally uses the GAN-based network to preserve the static characteristics of warped clothes, struggles to maintain the static characteristics of clothing well under high-resolution images.

\subsection{Quantitative Results}
The quantitative evaluation of the try-on task is challenging due to the absence of a real reference person with target clothes. In PLTON, we examine the paired and unpaired settings for image reconstruction and clothing item manipulation. For the unpaired setting, we utilize Fréchet Inception Distance (FID~\cite{FID}) as the evaluation metric, while for the paired setting, we use Learned Perceptual Image Patch Similarity (LPIPS~\cite{lpips}) as the reconstruction metric. However, it should be noted that the paired setting may not be suitable for virtual try-on, as discussed in ~\cite{PFAFN, RT-VTON}. Besides, the methods of pasting the warped clothes back to the generated image by learning the alpha mask can effectively improve the reconstruction metric. Our quantitative results on two test sets (TEST1 and TEST2) are given in Table ~\ref{tab:quantitative}. PLTON achieves state-of-the-art results by a large margin on the unpaired setting.
% In PLTON, we examine the paired and unpaired settings for image reconstruction and clothing item manipulation. For the unpaired setting, we utilize Fréchet Inception Distance (FID~\cite{FID}) as the evaluation metric, while for the paired setting, we use Learned Perceptual Image Patch Similarity (LPIPS~\cite{lpips}) as the reconstruction metric. However, it should be noted that paired setting may not be suitable for virtual try-on, as discussed in ~\cite{PFAFN, RT-VTON}. Our quantitative results, presented in Table ~\ref{tab:quantitative}, demonstrate that our method outperforms the baselines on all evaluation metrics at a resolution of 1024×768.

\begin{figure}[thb]
\vspace{-5pt}
\begin{center}
\includegraphics[width=0.95\linewidth]{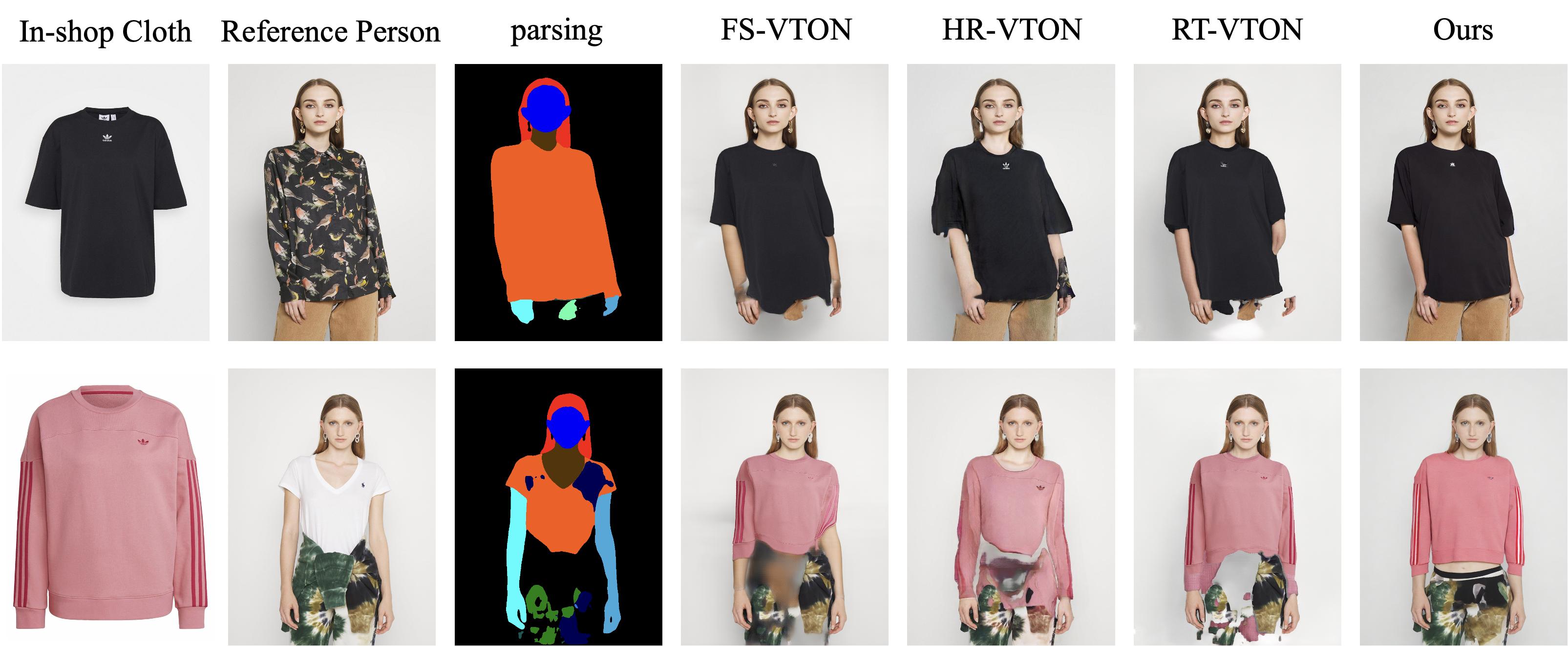}
\end{center}
   \caption{Visual comparison of traditional virtual try-on methods and ours. The case when the parsing of the reference person goes wrong is chosen to demonstrate the robustness of our method.}
\label{fig:robustness_seg}
\vspace{-10pt}
\end{figure}

\subsection{Robustness Analysis}
\begin{figure}[thb]
\vspace{-10pt}
\begin{center}
\includegraphics[width=0.95\linewidth]{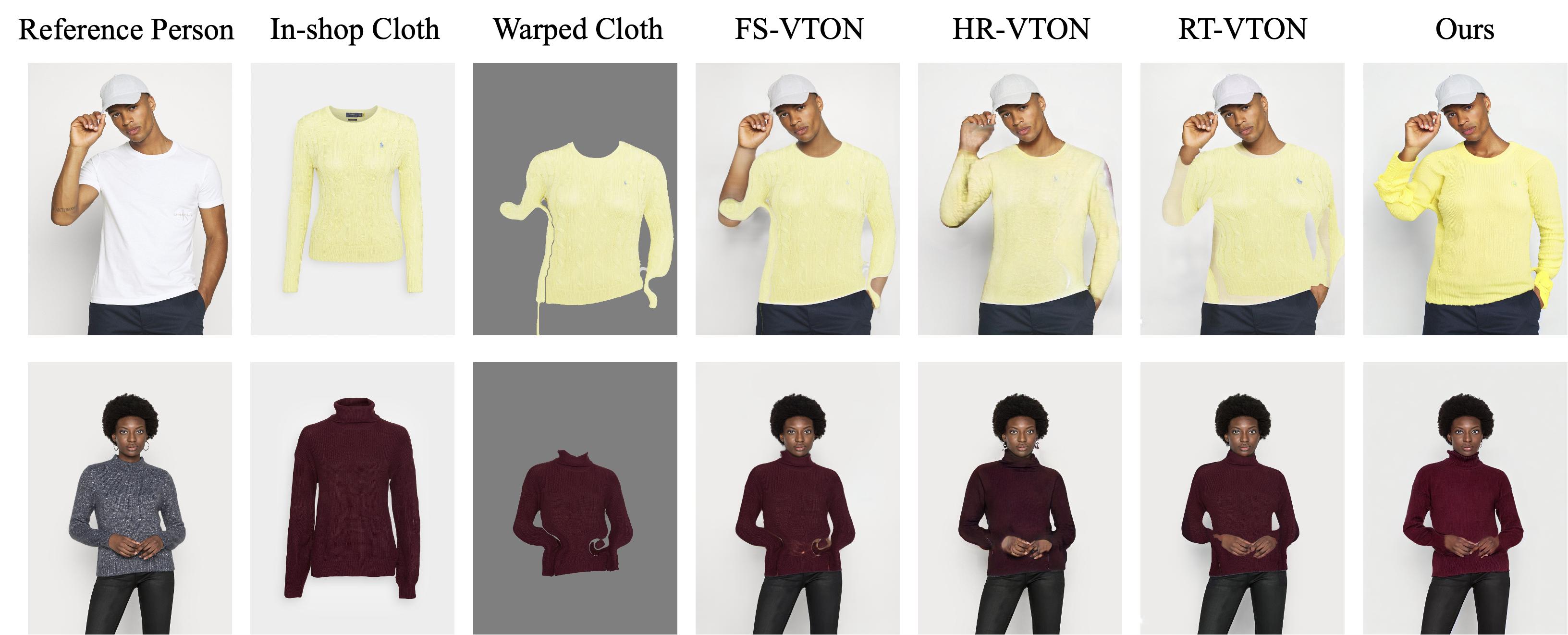}
\end{center}
   \caption{Visual comparison results between ours and traditional virtual try-on methods on slightly difficult (raised hands, crossed hands) model images.}
\label{fig:robustness_warp}
\vspace{-10pt}
\end{figure}

In the realm of virtual try-on algorithms, traditional methods have been categorized into flow-based and seg-based approaches. However, these methods often fall short when faced with inaccurate human parsing or complex poses, resulting in subpar outcomes. Our research has shown that PLTON exhibits greater robustness than baselines. For instance, in Figure~\ref{fig:robustness_seg}, when the parsing of reference person misses the trousers area, traditional methods tend to omit the trousers altogether, whereas PLTON is capable of complementing the missing piece. Similarly, in Figure~\ref{fig:robustness_warp}, when the model assumes a slightly complicated pose, such as crossed or raised hands, the warped clothes obtained by the warping module are often suboptimal, leading to inferior results. However, PLTON leverages FS-VTON badly warped clothes as guidance and can tolerate improper distortion errors, resulting in more realistic clothes details.

\begin{figure}[htb]
\begin{center}
\includegraphics[width=0.95\linewidth]{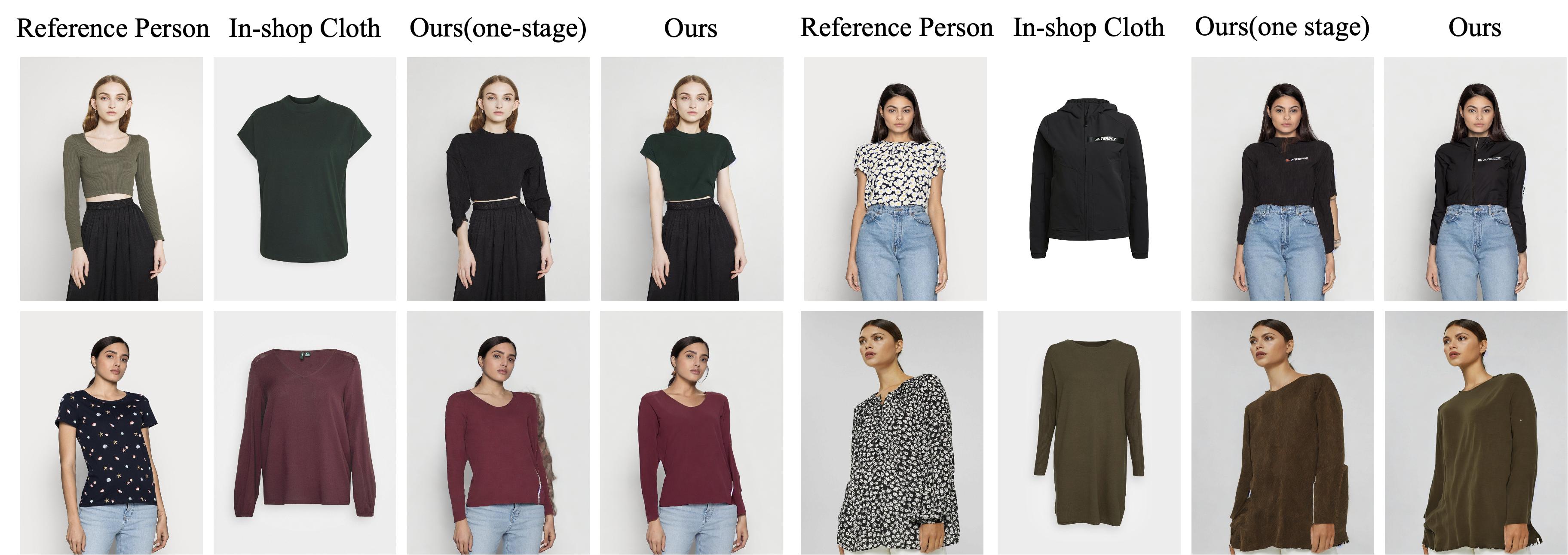}
\vspace{-10pt}
\end{center}
   \caption{Visual comparison between one-stage and two-stage inference methods in high-resolution parallax comparison results}
\label{fig:ablation1}
\vspace{-10pt}
\end{figure}

\begin{figure}[htb]
\begin{center}
\includegraphics[width=0.95\linewidth]{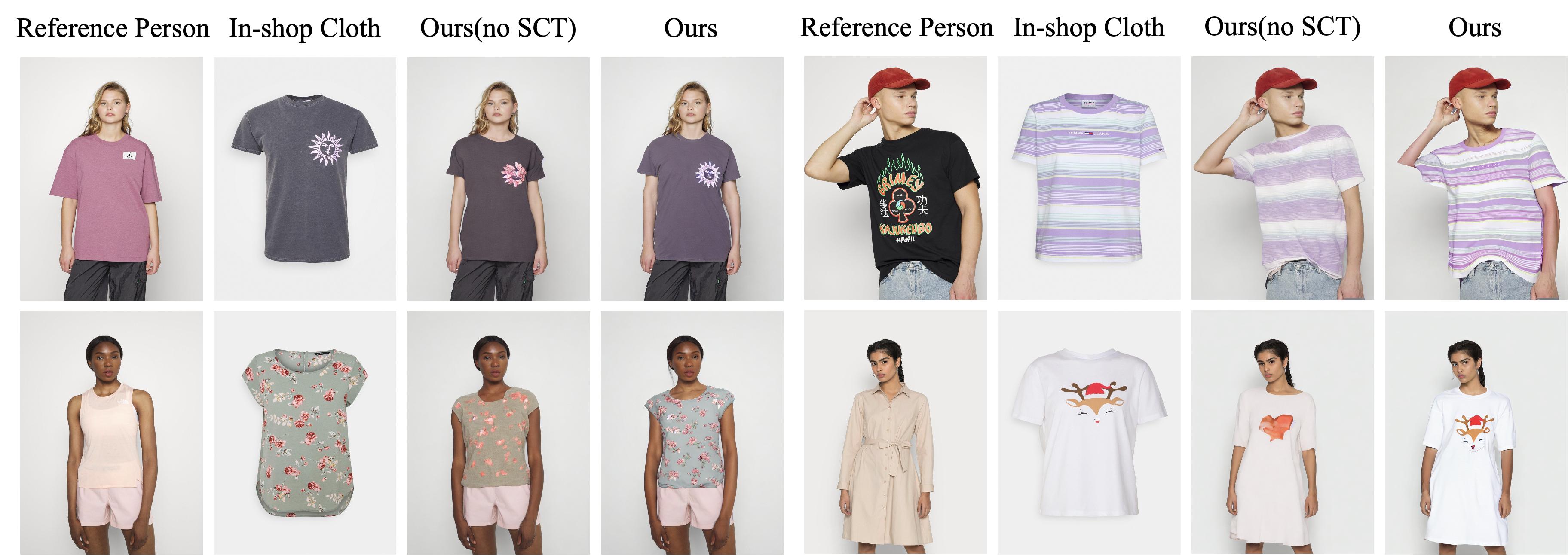}
\end{center}
   \caption{Visual ablation study of Static Extractor in PLTON.}
\label{fig:ablation2}
\vspace{-10pt}
\end{figure}

\begin{figure}[htb]
\vspace{-10pt}
\begin{center}
\includegraphics[width=0.85\linewidth]{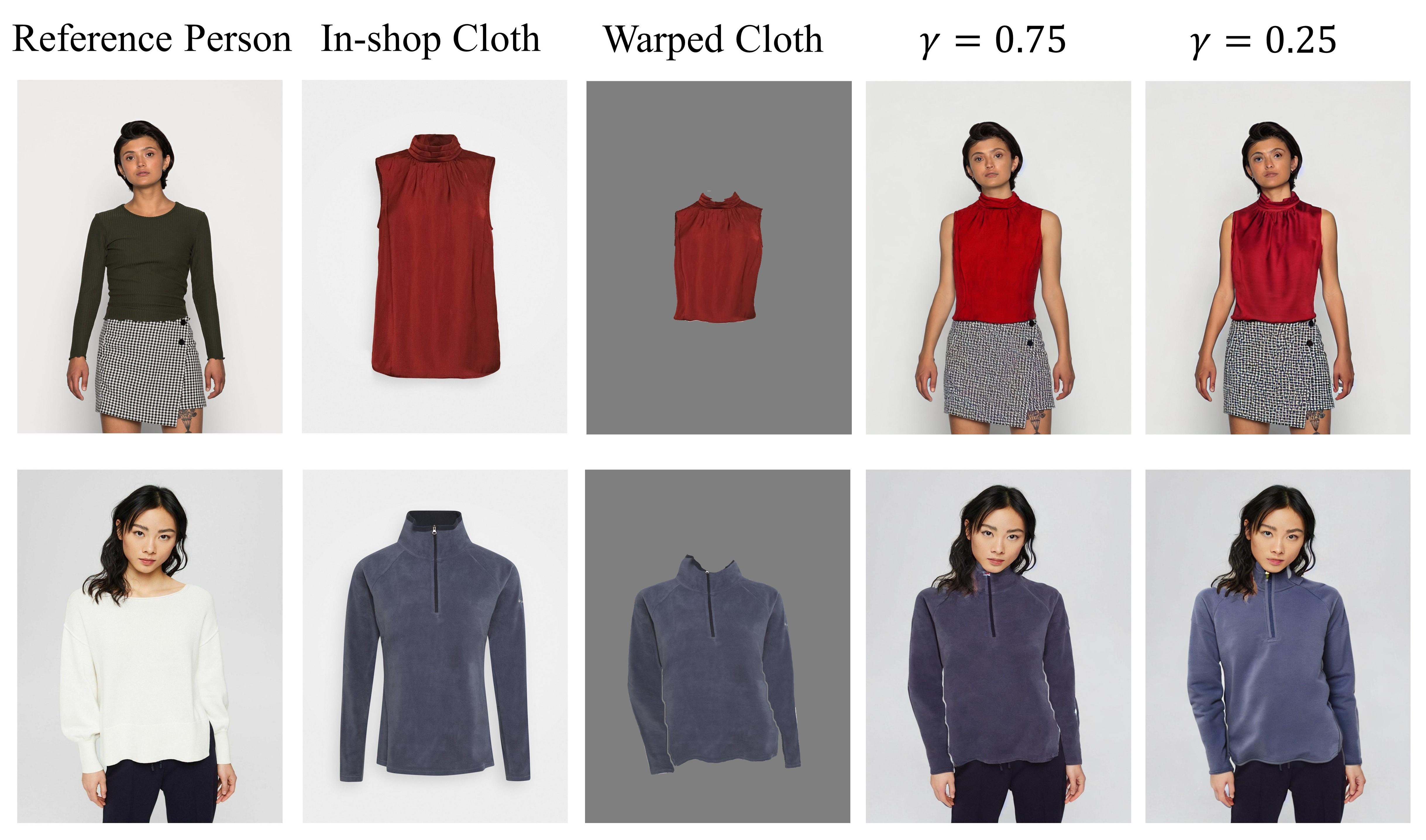}
\end{center}
   \caption{Visual comparsion of difference $\gamma$ in blended denoising.}
\label{fig:ablation3}
\vspace{-10pt}
\end{figure}

\subsection{Ablation Study}
Our ablation research mainly focuses on three aspects. Firstly, we investigate the effectiveness of Static Extractor, as an additional, conditional information control. Secondly, we explore the necessity of two-stage inference. Lastly, we analyze the influence of different parameters on clothes' static feature retention and dynamic feature generation.

\noindent \textbf{Effectiveness of two-stage inference.}\quad 
Data and computing resource constraints limit our training process to 512 × 512 resolution. However, we propose two solutions to obtain high-resolution images of 1024 during inference. The first is a single-stage inference, where high-resolution images are directly inferred. The second is a two-stage inference method, where we first infer a 768 × 576 image and then use it as guidance to generate high-resolution images. As illustrated in Figure~\ref{fig:ablation1}, direct high-resolution reasoning can lead to the problem of "repeat pattern", resulting in issues such as short-sleeved clothes becoming long-sleeved and sleeve ghosting. In virtual try-on, our two-stage inference method has proven effective in using the coarse image as a guide network to generate high-resolution images and avoid the problem of repeat patterns.

\noindent \textbf{Ablation on Static Characteristics Transformation.}\quad 
PLTON's Dual Feature Render module is based on stable diffusion. The module decouples the clothes features into dynamic and static features to simplify and speed up network training. The dynamic feature uses in-shop cloth with more information to replace the simple prompt, while the static feature uses the canny map of the distorted clothes as an additional condition to guide image generation. Figure~\ref{fig:ablation2} shows the effectiveness of the static feature transfer module as we compare it with the results obtained by reasoning when the static features are set to 0. The comparison reveals that without the guidance of static features, the network struggles to generate clothes textures and logos details with a small amount of data.

\noindent \textbf{Effectiveness of Blended Denoising.}\quad 
In the blended denoising process in PLTON, the parameter $\gamma$ plays a crucial role in controlling the reference degree of the network to the warped clothes. It guides the diffusion process toward the correct spatial layout and colors. However, it is essential to note that the value of $\gamma$ should not be too large as it can affect the network's ability to generate the clothes' dynamic features (e.g. shadow and folds) and its robustness. Figure~\ref{fig:ablation3} illustrates this point, where a value of 0.75 results in the network referring too much to the distorted clothes, leading to unnatural dynamic features and retention of defects.

\begin{table}[htb]
\begin{center}
  \caption{Preference comparison. A user study is given by the preference ratio for our method, which is the higher 705 the better.}
  \label{tab:user_study}
  \begin{tabular}{ccc}
  \hline
  \textbf{Method} & \textbf{TEST1} & \textbf{TEST2} \\
  \hline
  FS-VTON & 74.7\% & 73.7\% \\
  HR-VTON & 73.2\% & 70.2\% \\
  RT-VTON & 70.2\% & 75.8\% \\
  \hline
  % \bottomrule
\end{tabular}
\end{center}
\vspace{-10pt}
\end{table}

\subsection{User Study}
Image metrics may have limitations in depicting the try-on quality. In order to further demonstrate the superiority of our method, we found 24 volunteers to participate in our user study. Each volunteer is assigned 30 images, and each image contains a target cloth image, a reference person, a result from PLTON, and a result randomly selected from baselines. The user study results, as shown in ~\ref{tab:user_study}, clearly demonstrate that PLTON outperforms the existing state-of-the-art methods.

%% file: sections/conclusion.tex
\section{Conclusion}
% In this work we propose a Product-level Virtual try-on pipeline based on diffusion prior. Our key motivation is to divide the try-on into two schematics as static characteristics transformation and adaptive dynamic rendering in replace of the conventional split-transform-merge pipeline. The proposed dual feature renderer adaptively blends the deformed clothes in the denoising process and thus generate coherent clothes folds and shading with well-preserved details. Extensive experiments on high-resolution datasets demonstrate the superiority of our method especially on qualitative visual results.
In this work, we propose a product-level virtual try-on pipeline based on diffusion priors. Our main motivation is to split the fitting into two schematics, static feature transformation, and adaptive dynamic rendering, to replace the traditional split-transform-merge pipeline. The proposed dual-feature renderer adaptively blends deformed clothes during denoising, resulting in coherent clothes wrinkles and shadows with well-preserved details. Extensive experiments on high-resolution datasets demonstrate the superiority of our method, especially in qualitative visual results.